# Deblured Gaussian Blurred Images

Mr. Salem Saleh Al-amri[1], Dr. N.V. Kalyankar[2] and Dr. Khamitkar S.D [3]

**Abstract**—*This paper attempts to undertake the study of Restored Gaussian Blurred Images. by using four types of techniques of deblurring image as Wiener filter, Regularized filter ,Lucy Richardson deconvlutin algorithm and Blind deconvlution algorithm with an information of the Point Spread Function (PSF) corrupted blurred image with Different values of Size and Alfa and then corrupted by Gaussian noise. The same is applied to the remote sensing image and they are compared with one another, So as to choose the base technique for restored or deblurring image.This paper also attempts to undertake the study of restored Gaussian blurred image with no any information about the Point Spread Function (PSF) by using same four techniques after execute the guess of the PSF, the number of iterations and the weight threshold of it. To choose the base guesses for restored or deblurring image of this techniques.*

**Index Terms**—Blur/Types of Blur/PSF; Deblurring/Deblurring Methods.

——————————— ◆ ———————————

## 1 INTRODUCTION

The restoration image is very important process in the image processing to restor the image by using the image processing techniques to easily understand this image without any artifacts errors.In this case there are many studies undertaken in that scope and this some of these studies: In blind deconvolution, the goal is to deblur an image with (total of partial) lack of knowledge about the blurring operator to solve this problems he is proposed two alternative approaches to blind deconvolution: (i) simultaneously estimate the image and the blur (ii) perform a previous step of blur estimation and then feed this blur estimate to a classical non-blind image deblurring algorithm [1].The present is a novel algorithm to estimate direction and length of motion blur, using Radon transform and fuzzy set concepts. This method was tested on a wide range of different types of standard images that were degraded with different directions (between 0° and 180°) and motion lengths (between 10 and 50 pixels). The results showed that the method works highly satisfactory for SNR >22 dB and supports lower SNR compared with other algorithms [2].For correct restoration of the degraded image; it is useful to know the point-spread function (PSF) of the blurring system. We propose straightforward method to restore Gaussian blurred images given only the blurred image itself, the method first identifies the PSF of the blur and then use it to restore the blurred image with Standard restoration filters [3].The conventional Lucy-Richardson (LR) method is nonlinear and therefore its convergence is very slow. We present a novel method to accelerate the existing LR method by using an exponent on the correction ratio of LR; we present an adaptively accelerated Lucy-Richardson (AALR) method for the restoration of an image from its blurred and noisy version. That proposed AALR method shows better results in terms of low root mean square error (RMSE) and higher signal-to-noise ratio (SNR), in approximately 43% fewer iterations than those required for LR method [4].

## 2 BLURRING

Blur is unsharp image area caused by camera or subject movement, inaccurate focusing, or the use of an aperture that gives shallow depth of field. The Blur effects are filters that smooth transitions and decrease contrast by averaging the pixels next to hard edges of defined lines and areas where there are significant color transition.

### 2.1 Blurring Types
In digital image there are 3 common types of Blur effects:

### 2.1.1 Average Blur
The Average blur is one of several tools you can use to remove noise and specks in an image. Use it when noise is present over the entire image.
This type of blurring can be distribution in horizontal and vertical direction and can be circular averaging by radius R which evaluated by the formula:

$R = \sqrt{h^2 + v^2}$

Where: h is the horizontal size blurring direction and v is vertical blurring size direction is the radius size of the circular average blurring

### 2.1.2 Gaussian Blur
Gaussian Blur is that pixel weights aren't equal - they decrease from kernel center to edges according to a bell-shaped curve .The Gaussian Blur effect is a filter that blends a specific number of pixels incrementally, following a bell-shaped curve. The blurring is dense in the center and feathers at the edge. Apply Gaussian Blur to an image when you want more control over the Blur effect. Gaussian blur depends on the Size and Alfa.

### 2.1.3 Motion Blur
The Motion Blur effect is a filter that makes the image appear to be moving by adding a blur in a specific direction.
The motion can be controlled by angle or direction (0 to 360 degrees or –90 to +90) and/or by distance or intensity in pixels (0 to 999), based on the software used.

## 3 DEBLURRING

### 3.1 Deblurring Model
A blurred or degraded image can be approximately described by this equation:

$$g(x,y) = PSF * f(x,y) + \rho(x,y) \qquad (1)$$

Where: g the blurred image, PSF distortion operator called Point Spread Function, f the original true image and $\rho$ Additive noise, introduced during image acquisition, that corrupts the image

### 3.1.1 Point Spread Function (PSF)
Point Spread Function (PSF) is the degree to which an optical system blurs (spreads) a point of light. The PSF is the inverse Fourier transform of Optical Transfer Function (OTF).in the frequency domain ,the OTF describes the response of a linear, position-invariant system to an impulse.OTF is the Fourier transfer of the point (PSF).

### 3.2 Deblurring Methods
Our paper applys four methods of deblurring image:

### 3.2.1 Wiener Filter Deblurring Method
Wiener filter is a method of restoring image in the presence of blur and noise.The frequency-domain expression for the Wiener filter is:

$W(s) = H(s)/F^{+}(s)$, $H(s) = F_{x,s}(s)\, e^{as} / F^{-}_{x}(s) \qquad (2)$

Where: F(s) is blurred image, $F^+(s)$ causal, $F^-_x(s)$ anti-causal

### 3.2.2 Regularized Filter Deblurring Method
### 3.2.3 Lucy-Richardson Algorithm Method

The Richardson–Lucy algorithm, also known as Richardson–Lucy deconvolution, is an iterative procedure for recovering a latent image that has been the blurred by a known PSF.

$$C_i = \sum_j p_{ij} u_j \qquad (3)$$

Where

$P_{ij}$ is PSF at location i and j, $u_j$ is the pixel value at location j in blurred image.

$C_i$ is the observed value at pixel location i.

Iteration process to calculate $u_j$ given the observed $c_i$ and known $p_{ij}$

$$u_j^{(t+1)} = u_j^t \sum_i \frac{C_i}{c_i} p_{ij} \qquad (4)$$

Where

$$c_i = \sum_j u_j^{(t)} p_{ij} \qquad (5)$$

### 3.2.3 Blind Deconvolution Algorithm Method

Definition of the blind deblurring method can be expressed by:

g(x, y) =PSF * f(x,y) + η(x,y)     (6)

Where: g (x, y) is the observed image, PSF is Point Spread Function, f (x,y) is the constructed image and η (x,y) is the additive noise term .

## 4 EXPERIMENTS VERIFICATIONS
### 4.1 Testing Procedure

The deblurring was implemented using (MATLAB R2007a, 7.4a) and tested Gaussian blur type with different Size and Alfa with help of PSF function corrupted on the images illustrated in the Fig.1.

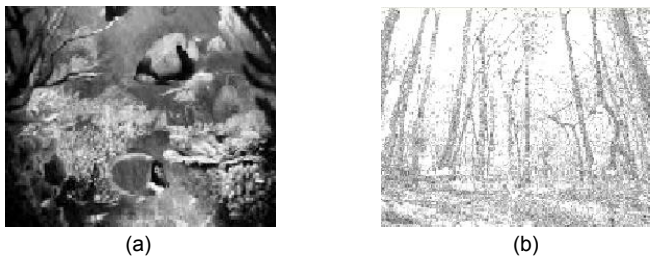

(a)                                         (b)
Fig.1. Original Image

Three types of deblurring methods are implemented: Wiener Filter Deblurring Method, Regularized Filter Deblurring Method, Lucy-Richardson Algorithm Method and blind Algorithm Method applying with PSF is known in the two cases:
(i) When noise is not added to the image (ii) with add noise to the image.
Also same four methods are applied with no information about PSF.

### 4.2 SIMULATION RESULTS

The performance results applied by two cases of the PSF function:

### 4.2.1 Deblurring with known PSF

The performance evaluations of the deblurring operation with known PSF can be implemented by two categories: the first category is a known amount of blur, but no noise, was added to an image, and second category is a known amount of blur and noise add to the image then the image was filtered to remove this known amount of blur and noise using Wiener, regularized and Lucy-Richardson,blind Algorithm deblurring methods. In the first category the regularized, Wiener and blind techniques produced what appeared to be the best results but it was surprising that the Lucy-Richardson technique produced the worst results in this instance see this result in the Fig.2.

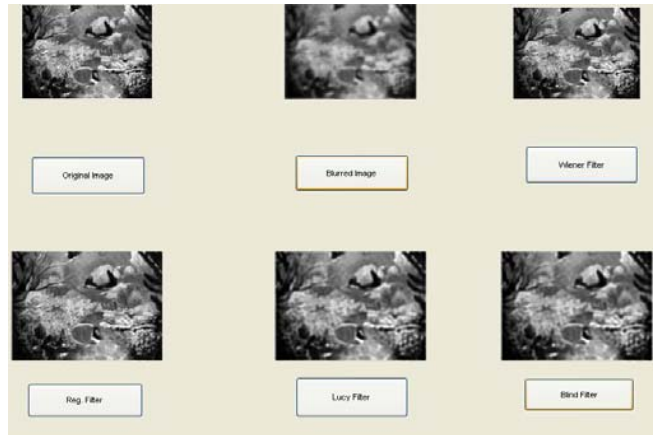

Fig.2. Deblurring image without add noise when PSF Known, image blurred by Gaussian blurred hsize =19 and Alfa= 0.3

In second category when Gaussian noise was added to the image in addition to blur the Lucy-Richardson algorithm actually performed the best results from the Wiener, Regularized and blind techniques. These results can be seeing in the Fig.3.

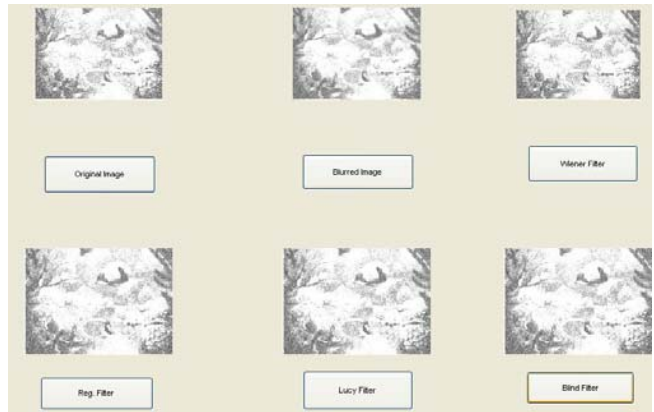

Fig.3. Original image with add Gaussian noise Alfa=0.5 when PSF Known

### 4.2.2 Deblurring with no PSF information

When no information about the original PSF the above techniques is not very useful techniques to Deblurring images. In this case applied another technique is called Blind deconvlution technique after executing the guess of the PSF, the number of iterations and the weight threshold of it. After much experimentation, it turned out that the weight threshold should be set between 0.10 and 0.25, the PSF matrix size should be set to 15x15, and the number of iterations should be any number more than 30.In this paper the best result is got when the PSF size is 19*19, iteration is 60 and weight threshold is 0.2 which illustrated in the Fig.4.

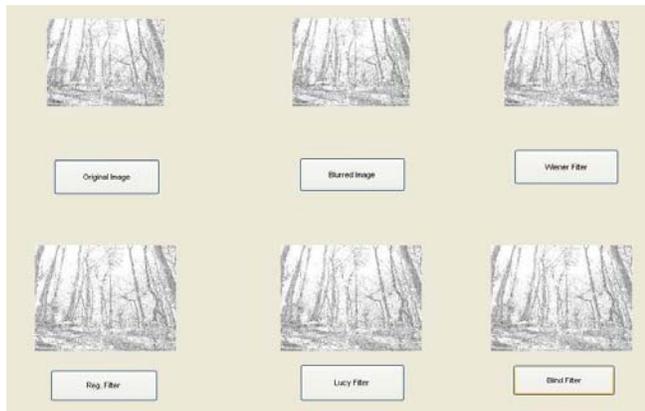

Fig.4. Deblurring image with no information of PSF

## 5 CONCLUSIONS

In this paper, the comparative studies undertaken two case: The first case with an information about PSF an second case with no information about PSF.In first case a comparative studie is explained & experiments are carried out for different techniques Wiener filter, regularized filter is the best techniques to deblurring of image sensing when don't noise in image see this in the Fig.2. But when noise is presented with blur the Lucy-Richardson algorithmic technique is the best techniques see in the Fig.3.Second case of the comparative study is explained & experiments are carried out for different techniques blind deconveluation algorithmic technique is the best techniques when the PSF size is 19*19, iteration is 60 and weight threshold is 0.2 which illustrated in the Fig.4 above.
.

## AUTHORS

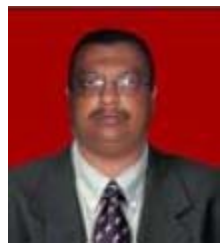

**Mr. Salem Saleh Al-amri.**Received the B.E degree in, Mechanical Engineering from University of Aden, Yemen, Aden in 1991, the M.Sc.degree in, Computer science (IT) from North Mahrashtra University (N.M.U), India, Jalgaon in 2006, Research student Ph.D in thedepartment of computer science (S.R.T.M.U), India, Nanded.

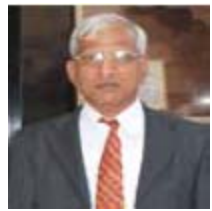

**Dr. N.V. Kalyankar**.B Sc.Maths, Physics, Chemistry, Marathwada University, Aurangabad, India, 1978. M Sc.Nuclear Physics,Marathwada University, Aurangabad,India, 1980.Diploma in Higher Education,Shivaji University, Kolhapur, India,1984.Ph.D. in Physics, Dr.B.A.M.University**,** Aurangabad, India,1995.Principal Yeshwant Mahavidyalaya College, Membership of Academic Bodies,Chairman, Information Technology Society State Level Organization, Life Member of Indian Laser Association**,** Member Indian Institute of Public Administration, New Delhi**,** Member Chinmay Education Society, Nanded.He has one publication book, seven journals papers, two seminars Papers and three conferences papers.

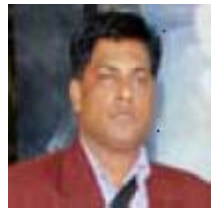

**Dr. S. D Khamitkar.** M. Sc. Ph.D.Computer Science Reader & Director(School of Computational Science)Swami Ramanand Teerth Marathwada University, Nanded**,**14 Years PG Teaching**,** Publications 08 International,Research Guide (10 Students registered),Member Board of Studies(Computer Application),Member Research and RecognitionCommittee (RRC) (Computer Studies).